\documentclass[sigconf]{acmart}

\copyrightyear{2026}
\acmYear{2026}
\setcopyright{cc}
\setcctype{by}
\acmConference[GeoSearch '26]{The 5th ACM SIGSPATIAL International Workshop on Searching and Mining Large Collections of Geospatial Data}{November 03--06, 2026}{Riverside, CA, USA}
\acmBooktitle{The 5th ACM SIGSPATIAL International Workshop on Searching and Mining Large Collections of Geospatial Data (GeoSearch '26), November 03--06, 2026, Riverside, CA, USA}
\acmDOI{10.1145/3849732.3857333}
\acmISBN{979-8-4007-3058-0/2026/11}
\usepackage{booktabs}
\usepackage{amsmath}
\usepackage{graphicx}
\newcommand{\conf}{\mathcal{C}}
\newcommand{\acc}{\mathrm{acc}}

\title{When Satellite and Street Views Disagree: Cross-View Complementarity in Disaster Damage Assessment}

\author{Yifan Yang}
\authornote{Both authors are corresponding authors.}
\email{yyang295@tamu.edu}
\affiliation{%
  \department{Department of Geography}
  \institution{Texas A\&M University}
  \city{College Station}
  \state{Texas}
  \country{USA}
}

\author{Lei Zou}
\authornotemark[1]
\email{lzou@tamu.edu}
\affiliation{%
  \department{Department of Geography}
  \institution{Texas A\&M University}
  \city{College Station}
  \state{Texas}
  \country{USA}
}

\begin{abstract}
Satellite and street-view imagery see different parts of a damaged
building: roofs from above, facades and debris from the ground. Average
accuracy hides whether fusion methods use this complementary evidence. We
study the \emph{conflict cases}, the samples on which independently trained
single-view models disagree, as a diagnostic subset. On paired collections
from the 2025 Eaton wildfire and Hurricanes Ian and Milton, conflicts make
up 9.6--33.2\% of the test data. Fusion improves conflict accuracy over
either view, yet an oracle that selects the correct single view per
conflict remains 0.19--0.32 above the best evaluated fusion method. A linear fusion gate over
building-visibility features, calibrated confidences, and disagreement
recovers part of this headroom where ground photographs are aimed at the
building: on Eaton it beats end-to-end fusion by $+0.072$ conflict accuracy
($p<10^{-4}$) and calibrated averaging by $+0.051$ ($p=0.0001$); on the
panoramic hurricane datasets it is indistinguishable from end-to-end fusion.
Cropping panoramas toward the building raises the conflict gain of fusion,
whereas random crops of the same size do not. The spatial density of
conflicts, computed without labels, correlates with tile-level wildfire
damage (Spearman $r=0.615$, 25 tiles). Conflict cases thus show how much
complementary evidence fusion uses, which view carries it under each hazard
and capture setting, and where current methods fall short.
\end{abstract}



\begin{document}

\maketitle

\begin{figure*}[t]
  \centering
  \includegraphics[width=0.44\textwidth]{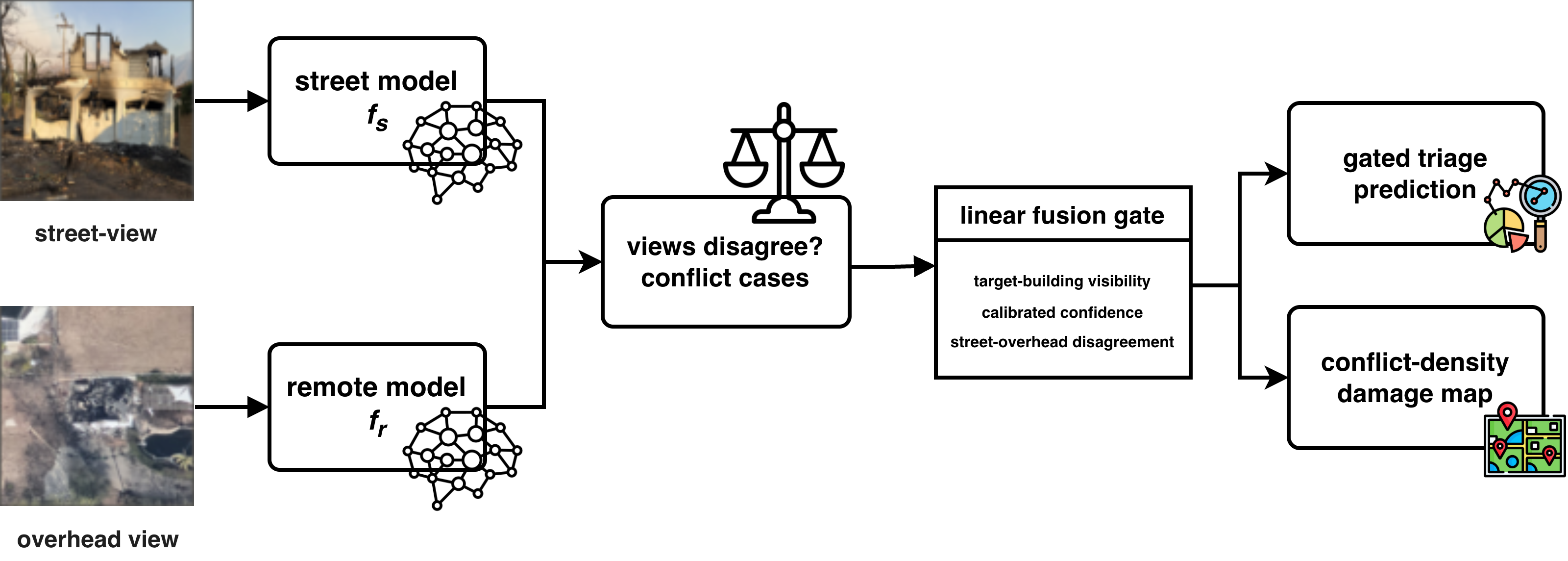}
  \Description{Pipeline diagram: a street-view image feeds a street model and
  an overhead image feeds a remote model; where the two disagree, conflict
  cases pass to a linear fusion gate over building visibility, calibrated
  confidence and street-overhead disagreement, which outputs a gated
  prediction; the spatial density of conflicts gives a damage map.}
  \caption{Overview. Single-view models define the conflict cases; a linear
  fusion gate decides how much weight each view receives per sample (the
  implemented gate also mixes in the end-to-end fusion model, omitted here);
  conflict density is a label-free damage indicator (Eaton-fire example).}
  \label{fig:pipeline}
\end{figure*}

\begin{table*}[t]
  \caption{Test-set results (five-seed means; accuracy std $\le 0.03$).
  Conflict accuracy is computed on each seed's conflict subset. Closure
  follows Eq.~\eqref{eq:closure} applied to the displayed means; the oracle
  reaches 0.961 / 0.896 / 0.964 conflict accuracy, i.e.\ 0.475 / 0.367 /
  0.407 above the better single view.}
  \label{tab:main}
  \centering
  \footnotesize
  \setlength{\tabcolsep}{5pt}
  \begin{tabular}{lccccccccc}
    \toprule
    & \multicolumn{3}{c}{Eaton wildfire (conflict rate 9.6\%)} & \multicolumn{3}{c}{Hurricane Ian (33.2\%)} & \multicolumn{3}{c}{Hurricane Milton (25.7\%)} \\
    \cmidrule(lr){2-4}\cmidrule(lr){5-7}\cmidrule(lr){8-10}
    Method & Acc. & Conflict acc. & Closure & Acc. & Conflict acc. & Closure & Acc. & Conflict acc. & Closure \\
    \midrule
    street\_only & 0.931 & 0.486 & --- & 0.703 & 0.529 & --- & 0.759 & 0.557 & --- \\
    remote\_only & 0.929 & 0.475 & --- & 0.648 & 0.367 & --- & 0.721 & 0.407 & --- \\
    concat & 0.930 & 0.674 & 0.40 & 0.709 & 0.588 & 0.16 & 0.766 & 0.622 & 0.16 \\
    crossview (end-to-end fusion) & 0.940 & 0.699 & 0.45 & \textbf{0.735} & \textbf{0.624} & \textbf{0.26} & 0.776 & \textbf{0.649} & \textbf{0.23} \\
    calibrated probability averaging & 0.954 & 0.732 & 0.52 & 0.731 & 0.617 & 0.24 & 0.776 & 0.626 & 0.17 \\
    \textbf{fusion gate (linear)} & \textbf{0.959} & \textbf{0.768} & \textbf{0.59} & 0.731 & 0.618 & 0.24 & \textbf{0.777} & 0.631 & 0.18 \\
    \bottomrule
  \end{tabular}
\end{table*}

\section{Introduction}
Building-level damage assessment after a disaster increasingly draws on two
image collections at once. Satellite imagery covers a whole region within
hours and is the basis of standardized damage
mapping~\cite{gupta2019creating}, but it mainly shows roofs. Ground-level
imagery, whether crowdsourced street-view panoramas or official inspection
photographs, shows the facades and debris that overhead sensors cannot
fully detect~\cite{zhai2020damage,yang2025hyperlocal}. The views are
complementary: a facade can be gutted under an unbroken roof. Methods that
pair them serve geolocalization and damage
mapping~\cite{li2025cross,yin2026triple,li2026towards}, multimodal
arbitration~\cite{yang2026damagearbiter}, and agentic assessment
pipelines~\cite{yang2026rapid,yang2026rapidmap}, and usually report that
fusion improves average accuracy over either view alone.

Average accuracy says little about whether this evidence is used, because
for most buildings the two views agree. The informative
records are the \emph{conflict cases}, the samples on which independently
trained single-view models disagree: at least one view carries evidence the
other lacks, whether damage is visible only from one side, vegetation hides
the building, or the views are misaligned. Conflicts are thus a focused
subset for asking how much complementary evidence fusion recovers, which
view carries it under a given hazard and capture setting, and how the views
can be combined.

This study examines cases in which damage assessments derived from
satellite imagery conflict with those derived from street-view imagery to
address four research questions. (1)~How much complementary headroom do
fusion methods leave on conflict cases? We compare them with an oracle that
selects whichever single view is correct. (2)~Can a simple, readable gate
recover part of it from cues about whether the ground image actually shows
the building? (3)~Does aiming the ground view at the building change the
value of fusion? We vary framing \emph{within} a dataset by controlled
cropping. (4)~Does the spatial density of conflicts,
which needs no labels once the models are trained, track damage
(Figure~\ref{fig:pipeline})? Code and protocols:
\url{https://github.com/rayford295/CrossViewGate}.

\section{Data and Protocol}
We use three paired street/overhead collections spanning two hazards and
two ground-view \emph{observation regimes}: property-centric
photographs aimed at a building, and panoramas that are not.

\textbf{Eaton wildfire (Altadena, CA; property-centric).} Building-level
CAL~FIRE Damage Inspection (DINS) photographs taken after the January 2025
fire, paired by location with post-event Very-High-Resolution (VHR) overhead tiles. The six-level
DINS scale is mapped to three ordinal classes (no/trace damage,
damaged--repairable, destroyed). The split has 6.5k/1,966/1,984
train/val/test pairs.
\textbf{Hurricane Ian (panoramic).} The CVIAN pairing released with
CVDisaster~\cite{li2025cross}: Mapillary 360$^\circ$ panoramas over Sanibel
Island, 27~cm post-event satellite imagery, and three-level damage-perception
labels, with 3,248/573/300 pairs.
\textbf{Hurricane Milton (panoramic).} A 360$^\circ$ street-view/satellite
pairing over Florida after the October 2024 hurricane, built in the same way;
1,740/307/254 pairs.

\paragraph{Splits and leakage.}
Eaton is split by property and tile, so all photographs of one structure fall
in one partition. For Ian, the headline numbers keep the released split for
comparability with CVDisaster. However, an audit against the official CVIAN
position file found that 299 out of the 300 test samples share a Mapillary
sequence with the training data, so we also evaluate a repaired protocol with
$0.005^\circ$ spatial blocks and a 25~m buffer. Under it, macro-F1 drops by
0.05--0.10 for every model, so index-based splits overstate spatial
generalization, although the single-view ranking (street above satellite) is
unchanged. The Ian results in Table~\ref{tab:main} should therefore be interpreted as
in-distribution results, not as the evidence of spatial transfer.

\section{Method}
\paragraph{Base models and conflict cases.}
For each dataset, we trained four models under one protocol:
\textit{street\_only} and \textit{remote\_only} (ResNet-18~\cite{he2016deep}
with a linear head), \textit{concat} (both encoders with an MLP head), and
\textit{crossview} (both encoders with a learned interaction head), which
serves as the end-to-end fusion reference. Training uses class-balanced
cross-entropy, 224-pixel inputs, AdamW, up to 15 epochs with early stopping
on validation macro-F1, and five seeds. Writing $s(x)$ and $r(x)$ for the two
single-view predictions, the \textbf{conflict subset} is
$\conf=\{x: s(x)\neq r(x)\}$. Conflict rates are 9.6\% on Eaton, 33.2\% on
Ian, and 25.7\% on Milton.

\paragraph{Oracle single-view gap.}
The oracle counts a conflict sample as correct if \emph{either} single view is
correct. For a method $m$ we report its conflict accuracy $\acc_{\conf}(m)$
and its \textbf{oracle-gap closure},
\begin{equation}
\mathrm{closure}(m)=
\frac{\acc_{\conf}(m)-\acc_{\conf}(\text{best single})}
     {\acc_{\conf}(\text{oracle})-\acc_{\conf}(\text{best single})},
\label{eq:closure}
\end{equation}
where "best single'' is the single-view model with the higher mean conflict
accuracy. The oracle only chooses \emph{which} of two fixed models to use,
so it measures the headroom of hard view selection; it is not an upper bound
for probability mixtures, which can in principle exceed it.

\paragraph{Calibration.}
Since miscalibration can distort averaging~\cite{guo2017calibration},
we fit one temperature per view on validation predictions and evaluated every
training-free fusion baseline on calibrated probabilities; the argmax, and
hence the conflict subset, is unchanged.

\paragraph{Visibility-conditioned fusion gate.}
A frozen SegFormer-B0~\cite{xie2021segformer} trained on ADE20K gives four
building-visibility features per street image: the building pixel ratio, the
same ratio inside a centered crop of half the image size, their difference (a
centering signal), and the normalized distance of the building centroid from
the image center. These features describe visible built structures in
general, not the specific target building. Seven more features describe the
predictions themselves: calibrated confidence and entropy for each view, the
confidence gap, the Jensen--Shannon divergence between the two calibrated
distributions, and a disagreement flag. Given the eleven z-scored features
$z$, the gate is a mixture of experts~\cite{jacobs1991adaptive} with linear
logits,
\begin{equation}
p(y\mid x)=\sum_{k}w_k(z)\,p_k(y\mid x),\qquad w(z)=\mathrm{softmax}(Wz+b),
\label{eq:gate}
\end{equation}
over $k\in\{\text{street},\text{remote},\text{crossview}\}$, fit on the
validation split by minimizing the negative log-likelihood of the mixture.
A \emph{two-expert} variant (street and remote only) has a single logit, so
each coefficient reads directly as a shift of weight toward one view. A small
MLP in place of the linear logits did not help on Eaton.

\paragraph{Field-of-view intervention.}
Each $1024\times512$ panorama is cropped to a $256\times256$ window with a
90$^\circ$ horizontal field of view on the horizon band. In the
\textbf{building-centered} variant the window is aimed at the circular-mean
azimuth of the SegFormer building mask. In the \textbf{random} variant it is
aimed at a reproducible uniformly random azimuth. The variants differ only
in where the window points. We retrained \textit{street\_only} and \textit{crossview} on each
variant (three seeds, three-epoch schedule, also used for the
original-panorama reference), reuse the remote models, and compare the
gain of fusion on conflicts.

\paragraph{Conflict density.}
We averaged test-set predictions over seeds, aggregate them to geographic
tiles, and computed per tile the hard conflict density (fraction of samples on
which the two argmax predictions disagree) and, as a control, the mean
single-view predictive entropy. Both are rank-correlated (Spearman) with the
tile's mean ordinal damage label.

\section{Results}
\label{sec:results}
\paragraph{Overall results and the oracle gap.}
For paired comparisons, we averaged each test sample's correctness difference
over the conflict seeds and applied a two-sided sign-flip
permutation test (20,000 permutations). Pooled differences weight samples
rather than seeds, so they differ slightly from Table~\ref{tab:main}. First,
end-to-end fusion beats the street model on conflicts on every dataset
(Eaton $+0.246$, $p<10^{-4}$; Ian $+0.078$, $p=0.013$; Milton $+0.141$,
$p=0.006$). Second, once the single-view models are converged and calibrated,
probability averaging is a strong baseline: on Eaton it exceeds end-to-end
fusion in both overall accuracy (0.954 vs.\ 0.940) and conflict accuracy
(0.732 vs.\ 0.699). Third, the
oracle headroom dwarfs these differences: picking the right single view per
sample would add 0.475 / 0.367 / 0.407 conflict accuracy over the better
single view, and still 0.193 / 0.272 / 0.315 over the best evaluated fusion
method.

\begin{figure}[t]
  \centering
  \includegraphics[width=0.85\columnwidth]{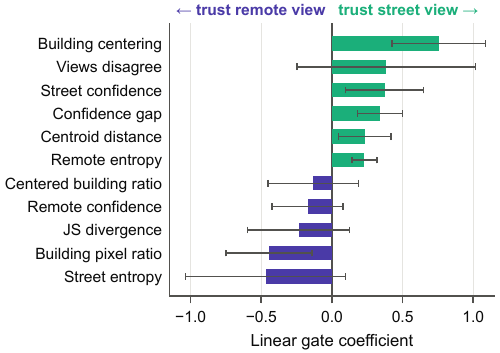}
  \Description{Horizontal bar chart of the eleven coefficients of the
  two-expert linear gate on the Eaton wildfire, with error bars over five
  seeds. Building centering has the largest positive coefficient; building
  pixel ratio and street entropy are negative; several error bars cross
  zero.}
  \caption{Coefficients of the two-expert (street vs.\ remote) linear gate on
  Eaton (mean $\pm$ std over five seeds). Positive values shift weight toward
  the street view; bars crossing zero have no stable sign across seeds.}
  \label{fig:gate}
\end{figure}

\paragraph{The fusion gate.}
On the property-centric dataset the gate gains $+0.072$ conflict accuracy over
end-to-end fusion ($p<10^{-4}$) and $+0.018$ on the full test set
($p<10^{-4}$); against calibrated averaging, the strongest baseline, it gains
$+0.051$ on conflicts (95\% bootstrap CI $[+0.026, +0.077]$, $p=0.0001$). Its
closure of 0.59 is the highest in Table~\ref{tab:main}. On the panoramic
datasets the street views rarely show a building prominently (mean building
pixel ratio 0.027 on Ian vs.\ 0.119 on Eaton), so the visibility features
carry little signal, and the gate is statistically indistinguishable from
end-to-end fusion (pooled differences $+0.005$, $p=0.70$ on Ian; $-0.015$,
$p=0.57$ on Milton). 
How much of the Eaton gain comes from visibility-based weighting rather than
from adding a third expert? The two-expert gate, which can only reweight
street and remote, reaches 0.734 conflict accuracy, on par with calibrated
averaging (0.732) and above end-to-end fusion ($+0.034$, $p=0.047$). Adding
crossview as a third expert lifts this to 0.768, so the headline gain combines
both effects. The two-expert coefficients (Figure~\ref{fig:gate}) are the
interpretable part. The largest weight is building centering ($+0.75\pm0.33$),
followed by the confidence gap and street confidence, while the raw building
pixel ratio is negative ($-0.45\pm0.31$). Weight on the street view thus rises
when the photograph is confident and the building sits in the middle of the
frame, not merely when much of it is built-up area.

\begin{table}[t]
  \caption{Field-of-view intervention (three seeds, three-epoch schedule;
  remote models unchanged). Conflict gain = crossview $-$ best single view on
  each variant's own conflict subset.}
  \label{tab:fov}
  \centering
  \scriptsize
  \setlength{\tabcolsep}{4pt}
  \begin{tabular}{llcc}
    \toprule
    Dataset & Street-view variant & Street acc. & Conflict gain \\
    \midrule
    Ian    & original panorama       & 0.709 & $+0.064$ \\
    Ian    & building-centered crop  & 0.648 & $\mathbf{+0.109}$ \\
    Ian    & random crop (control)   & 0.574 & $+0.042$ \\
    \midrule
    Milton & original panorama       & 0.740 & $+0.067$ \\
    Milton & building-centered crop  & 0.693 & $\mathbf{+0.149}$ \\
    Milton & random crop (control)   & 0.654 & $+0.055$ \\
    \bottomrule
  \end{tabular}
\end{table}

\begin{figure*}[t]
  \centering
  \includegraphics[width=0.62\textwidth]{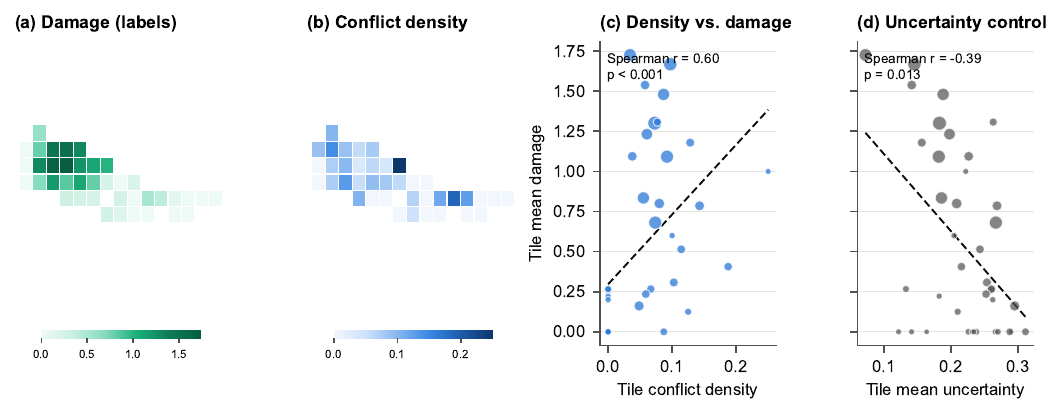}
  \Description{Four panels for the Eaton wildfire on a 0.01-degree grid: a tile
  map of labeled damage, a tile map of conflict density computed without
  labels, a scatter of density against damage with positive rank correlation,
  and the single-view uncertainty control with negative correlation.}
  \caption{Conflict density as a label-free damage indicator (Eaton wildfire,
  $0.01^\circ$ grid). (a)~Labeled tile damage. (b)~Tile conflict density
  from model predictions. (c)~Their rank correlation. (d)~The single-view
  uncertainty control, which anti-correlates with damage. Point size
  reflects tile sample count; $p$-values ignore spatial dependence.}
  \label{fig:density}
\end{figure*}

\paragraph{Field-of-view intervention.}
Building-centered cropping raises the conflict gain of fusion on both
hurricanes (Table~\ref{tab:fov}): from 0.064 to 0.109 on Ian and from 0.067 to
0.149 on Milton. Random crops with identical geometry do not (0.042 and
0.055), and the building-minus-random contrast is positive in all six
dataset--seed pairs. Cropping \emph{lowers} street-only accuracy yet \emph{raises}
the benefit of fusion, so the gain does not come from a stronger street
model, and the random control argues against a cropping artifact. This
supports building-oriented framing as a driver of the regime effect,
although a fixed-sample test with target-building annotations would be
stronger.

\paragraph{Conflict density as a spatial damage signal.}
On the wildfire, the conflict density of an overhead tile tracks its damage
(Spearman $r=0.615$, $p=0.001$ over 25 tiles; $r=0.60$ on the $0.01^\circ$
grid of Figure~\ref{fig:density}). The uncertainty control behaves very
differently: it \emph{anti-correlates} with damage ($r=-0.37$ to $-0.52$),
because parcels burned to the ground are easy, confident classifications.
Damage on
Eaton is spatially autocorrelated (Moran's $I=0.27$),
so the nominal $p$-values overstate the evidence. On Milton the signal is
weaker and not significant once the models converge ($r=0.289$, $p=0.14$),
because better single-view models disagree less.

\section{Discussion and Conclusion}
Conflict cases make the value of complementary views measurable. The oracle
headroom, a restricted choice between two existing models, exceeds every
fusion gain we observed, so current fusion leaves most view-specific
evidence unused. Which view carries it depends on hazard and capture
setting. On property-centric wildfire photographs, visibility cues identify
when the street view is decisive, and the gate, one frozen segmentation
pass plus a linear fit, is cheap and auditable: weight the street view when
it is confident and the building is centered. On
panoramic hurricane imagery the building is rarely prominent, and
calibrated averaging or end-to-end fusion suffices, and framing ground
photographs on structures raises the value of fusion.

\paragraph{Limitations.}
Hazard and capture regime are confounded across datasets, and the
property-centric regime rests on one wildfire. Milton is small (254 test
pairs); the released Ian split has sequence overlap and the repaired
protocol only six test blocks. The permutation tests ignore dependence
between nearby properties. We measure prediction complementarity, not which
component was damaged; conflicts may also stem from occlusion,
misalignment, or label ambiguity. The density signal needs spatially aware inference and
replication.

\paragraph{Conclusion.}
Where satellite and street views disagree, symmetric fusion leaves
substantial complementary evidence unused. A visibility-conditioned gate
recovers part of it for building-aimed photographs, and spatially
aggregated disagreement is a label-free damage indicator.
Mining conflicts turns ``does fusion help?'' into ``which view carries the
evidence, where, and why?''.

\begin{acks}
Supported by the U.S.\ National Science Foundation (Award 2318206) and the
Gulf Research Program of the National Academies of Sciences, Engineering,
and Medicine (SCON-10000653 and SCON-10001536).
Claude (Anthropic) and OpenAI Codex were used for language editing and for
checking reported numbers against the result logs. The authors take full
responsibility for the content.
\end{acks}

\bibliographystyle{ACM-Reference-Format}
\bibliography{references}

@inproceedings{gupta2019creating,
  title = {Creating xBD: A Dataset for Assessing Building Damage from Satellite Imagery},
  author = {Gupta, Ritwik and Goodman, Bryce and Patel, Nirav and Hosfelt, Ricky and Sajeev, Sandra and Heim, Eric and Doshi, Jigar and Lucas, Keane and Choset, Howie and Gaston, Matthew},
  booktitle = {Proceedings of the IEEE/CVF Conference on Computer Vision and Pattern Recognition Workshops},
  pages = {10--17},
  year = {2019}
}

@article{zhai2020damage,
  title = {Damage Assessment Using Google Street View: Evidence from Hurricane Michael in Mexico Beach, Florida},
  author = {Zhai, Wei and Peng, Zhong-Ren},
  journal = {Applied Geography},
  volume = {123},
  pages = {102252},
  year = {2020},
  publisher = {Elsevier}
}

@article{yang2025hyperlocal,
  title = {Hyperlocal Disaster Damage Assessment Using Bi-Temporal Street-View Imagery and Pre-Trained Vision Models},
  author = {Yang, Yifan and Zou, Lei and Zhou, Bing and Li, Daoyang and Lin, Binbin and Abedin, Joynal and Yang, Mingzheng},
  journal = {Computers, Environment and Urban Systems},
  volume = {121},
  pages = {102335},
  year = {2025},
  doi = {10.1016/j.compenvurbsys.2025.102335}
}

@article{li2025cross,
  title = {Cross-View Geolocalization and Disaster Mapping with Street-View and VHR Satellite Imagery: A Case Study of Hurricane IAN},
  author = {Li, Hao and Deuser, Fabian and Yin, Wenping and Luo, Xuanshu and Walther, Paul and Mai, Gengchen and Huang, Wei and Werner, Martin},
  journal = {ISPRS Journal of Photogrammetry and Remote Sensing},
  volume = {220},
  pages = {841--854},
  year = {2025},
  publisher = {Elsevier}
}

@article{yin2026triple,
  title = {Triple-Objective Cross-View Geolocalization of Disaster-Related VGI: The Case of Hurricane Ian},
  author = {Yin, Wenping and Deuser, Fabian and Liu, Ziqi and Wei, Jiaqi and Luo, Xuanshu and Werner, Martin and Li, Hao and Xue, Yong},
  journal = {International Journal of Geographical Information Science},
  volume = {40},
  number = {1},
  pages = {194--216},
  year = {2026},
  publisher = {Taylor \& Francis}
}

@article{li2026towards,
  title = {Towards Generative Location Awareness for Disaster Response: A Probabilistic Cross-View Geolocalization Approach},
  author = {Li, Hao and Deuser, Fabian and Yin, Wenping and Knoblauch, Steffen and Zhao, Wufan and Biljecki, Filip and Xue, Yong and Huang, Wei},
  journal = {ISPRS Journal of Photogrammetry and Remote Sensing},
  volume = {237},
  pages = {130--145},
  year = {2026},
  publisher = {Elsevier}
}

@article{yang2026damagearbiter,
  title = {DamageArbiter: A CLIP-Enhanced Multimodal Arbitration Framework for Hurricane Damage Assessment from Street-View Imagery},
  author = {Yang, Yifan and Zou, Lei and Gong, Wenjing and Fu, Kani and Li, Zongrong and Wang, Siqin and Zhou, Bing and Cai, Heng and Tian, Hao},
  journal = {arXiv preprint arXiv:2603.14837},
  year = {2026}
}

@inproceedings{he2016deep,
  title = {Deep Residual Learning for Image Recognition},
  author = {He, Kaiming and Zhang, Xiangyu and Ren, Shaoqing and Sun, Jian},
  booktitle = {Proceedings of the IEEE Conference on Computer Vision and Pattern Recognition},
  pages = {770--778},
  year = {2016}
}

@inproceedings{guo2017calibration,
  title = {On Calibration of Modern Neural Networks},
  author = {Guo, Chuan and Pleiss, Geoff and Sun, Yu and Weinberger, Kilian Q.},
  booktitle = {Proceedings of the 34th International Conference on Machine Learning},
  pages = {1321--1330},
  year = {2017},
  organization = {PMLR}
}

@inproceedings{xie2021segformer,
  title = {SegFormer: Simple and Efficient Design for Semantic Segmentation with Transformers},
  author = {Xie, Enze and Wang, Wenhai and Yu, Zhiding and Anandkumar, Anima and Alvarez, Jose M and Luo, Ping},
  booktitle = {Advances in Neural Information Processing Systems},
  volume = {34},
  pages = {12077--12090},
  year = {2021}
}

@article{jacobs1991adaptive,
  title = {Adaptive Mixtures of Local Experts},
  author = {Jacobs, Robert A. and Jordan, Michael I. and Nowlan, Steven J. and Hinton, Geoffrey E.},
  journal = {Neural Computation},
  volume = {3},
  number = {1},
  pages = {79--87},
  year = {1991}
}

@article{yang2026rapid,
  title = {{RAPID}: A Reproducible Multi-Agent Pipeline for Interpretable Disaster Damage Assessment from Satellite and Street-View Imagery},
  author = {Yang, Yifan and Gong, Wenjing and Zhang, Kaili and Zou, Lei and Tu, Zhengzhong and Li, Hao and Li, Zongrong and Ye, Xinyue},
  journal = {arXiv preprint arXiv:2606.21819},
  year = {2026}
}

@article{yang2026rapidmap,
  title = {{RAPIDMap}: Rapid Multi-Agent Pipeline for Interpretable Disaster Mapping from Satellite and Street-view Imagery},
  author = {Yang, Yifan and Zou, Lei},
  journal = {arXiv preprint arXiv:2609.00046},
  year = {2026}
}

\end{document}